# Some Advances in Role Discovery in Graphs

Sean Gilpin
*University of California, Davis*
`sagilpin@ucdavis.edu`

Chia-Tung Kuo
*University of California, Davis*
`tomkuo@ucdavis.edu`

Tina Eliassi-Rad
*Rutgers University*
`eliassi@cs.rutgers.edu`

Ian Davidson
*University of California, Davis*
`davidson@cs.ucdavis.edu`

September 8, 2016

## Abstract

Role discovery in graphs is an emerging area that allows analysis of complex graphs in an intuitive way. In contrast to other graph problems such as community discovery, which finds groups of highly connected nodes, the role discovery problem finds groups of nodes that share similar graph topological structure. However, existing work so far has two severe limitations that prevent its use in some domains. Firstly, it is completely unsupervised which is undesirable for a number of reasons. Secondly, most work is limited to a single relational graph. We address both these limitations in an intuitive and easy to implement alternating least squares framework. Our framework allows convex constraints to be placed on the role discovery problem which can provide useful supervision. In particular we explore supervision to enforce i) sparsity, ii) diversity and iii) alternativeness. We then show how to lift this work for multi-relational graphs. A natural representation of a multi-relational graph is an order 3 tensor (rather than a matrix) and that a Tucker decomposition allows us to find complex interactions between collections of entities (*E-groups*) and the roles they play for a combination of relations (*R-groups*). Existing Tucker decomposition methods in tensor toolboxes are not suited for our purpose, so we create our own algorithm that we demonstrate is pragmatically useful.

## 1  Introduction

Role discovery is a developing area that allows the simplification of graphs in a user-interpretable way. Consider a graph of $n$ nodes specified in an adjacency matrix **A**. Earlier efforts convert this matrix into a new $n \times f$ matrix **V** so that each node in the graph has a list of $f$ features [22]. Role discovery is then the computation of converting $V$ so that each node/user is mapped to a combination of roles (denoted by the $n \times r$ matrix **G**) and each role is defined with respect to the $f$ features (denoted by the $r \times f$ matrix **F**). This is accomplished by performing a non-negative matrix factor decomposition as shown below:

$$\underset{\mathbf{G},\mathbf{F}}{\operatorname{argmin}} \quad ||\mathbf{V} - \mathbf{GF}||_2 \qquad (1)$$
$$\text{subject to:} \quad \mathbf{G} \geq \mathbf{0}, \mathbf{F} \geq 0$$

The $n \times r$ matrix **G** when read row-wise indicates which of the $r$ roles each node plays and to what degree. The $r \times f$ matrix **F** when read row-wise defines each of the $r$ roles with respect to the $f$ features. The entries in **G** and **F** are non-negative real numbers signifying that each node can play each role to varying degrees and that different features define a role in varying degrees. This simplification of graphs into roles is not only intuitive for a domain expert, but it has been shown to be useful in a number of interesting settings including prediction, transfer learning, and sense making [21].

**Limitations in Existing Work.** However, all work developed so far has two limitations. Firstly, role discovery has been typically completely unsupervised in that the domain expert cannot easily inject their expertise and expectations into the simplification and secondly role discovery is typically performed on a single relational graph. We now discuss each limitation in turn.

Consider a domain expert that is looking for the simplest explanation of a graph during their exploratory phase of analysis. Existing work cannot specify how to emphasize this simplicity apart from requiring a small number of roles to be used. Other forms of parsimonious guidance such as requiring a node only be assigned to a few roles or making each role defined by only a small set of features is desirable but currently not possible. Similarly, if a decomposition yields a set of roles that are not actionable, not interesting or already known, the domain expert cannot enforce an *alternative* set of roles. These two recent trends in data mining – exploring the addition of positive and negative guidance – have been shown to have wide-scale application in the data mining literature [5][36]; but to our knowledge have not been applied to role discovery. Hence this work marks the first paper exploring guided role discovery.

To our knowledge previous work in role discovery only focuses on simple graphs with a single relational type. Conversely, many datasets are either directly multi-relational or can be modeled as a multi-relational graph. Consider an email graph, modeling just one relation `sent-mail-to`. This graph greatly masks the complexity of the underlying behavior occurring in the network. Instead many more insights could be found if say the *topic* of the email were also considered producing a multi-relational graph `sent-mail-to-y-about-x`. Similarly, consider a node-attributed social network graph, that is, each node has multiple labels. Such a graph can augment the basic `friend` relation by creating multiple relations such as `female-friend`, `school-friend` or `nearby-friend` by placing an edge between nodes that are friends that also share label values.

**Challenges.**

The challenge with adding guidance to role discovery is how to do so whilst still yielding an efficiently solvable algorithm. Pre-processing the graph or post-processing the results is undesirable, instead it is preferable to inject the guidance into the underlying algorithm that finds the roles. The alternating least square (ALS) is a popular and well understood algorithm for non-negative matrix factorization (NMF) for role discovery and the challenge is to add in guidance into this algorithm.

The challenge of role discovery in multi-relational graphs is two fold, the first is representational and the second is algorithmic.

**Representational Challenges.** How should a multi-relational graph be represented for effective and efficient role discovery? In Figure 6, we show how an order 3 tensor can compactly represent such a graph. Here, the first mode represents the entities (i.e., nodes) in the graph; the second mode is the features of each entity (obtained from our ReFeX package [22]), and the last mode represents the relations. The existing work on single-relation graphs uses non-negative matrix factorization. The analog PARAFAC (parallel factor) tensor decomposition for our multi-relational graph tensor, has several serious limitations. In particular, it requires each group of nodes to play exactly one role for exactly one set of relations. This is not due to the rank one decomposition assumption, but rather due to the simplified form of decomposition. This cardinality limitations greatly limits what can be found. Consider our aforementioned example of an email network, we could perhaps find that a group of people play the role of a `broker` for a particular email topic, `office-party`. Though useful, if those people also play a different role for the same email topic a PARAFAC decomposition could not find it. Similarly and most importantly, if another group were to play the role of a `peripheral figure` for the exact same topic (`office-party`) PARAFAC would not be able to discover this relation. It is precisely these types of complex multi-way interactions between people, roles and relations that we wish to discover. Hence we do not consider PARAFAC decompositions, though it would be the natural extension of our earlier work on role discovery to multi-relational graphs. Instead we use a Tucker decomposition shown in Figure 7 whose addition of a core tensor to the decomposition allows multiple groups of entities (E-groups) to play multiple roles for multiple groups of relations (R-groups). This allows very complex insights into the behavior in the graph to be found, but the challenge of how to interpret and use this core is critical to our work.

**Algorithmic Challenges.** The second challenge is that existing Tucker decompositions found in the popular Kolda Tensor Toolbox and Bro NWay Toolbox are not suited for our purpose. Existing toolboxes implement an orthogonality constraint on the factor matrices which in our context (where the tensor only contains non-negative values) means each group of entities must be distinct (i.e., non-overlapping) from every other group, and the same for roles and groups of relations. Similarly existing toolboxes typically do not enforce a non-negativity constraint on the core of the Tucker meaning if we use them we would have entities playing negative roles which does not make intuitive sense. Hence to better fit our needs of having interpretable insights on overlapping groups of entities (a.k.a. **E-groups**), roles, and groups of relations (a.k.a. **R-groups**), we develop our own algorithm, Multi-relational Role Discovery (MRD), shown in Algorithm 1.

Our work makes several contributions to the field of role discovery in graphs. With respect to guided role discovery we show:

- We provide a framework to encode guidance as a series of convex optimization problems each of which can be efficiently solved by our alternating least squares (ALS) algorithm. All data sets and code will be made available once the paper is accepted.

- Within our framework we explore guidance in the form of sparsity, diversity and orthogonality/alternativeness but other types of guidance are possible.

- We show that sparsity and diversity yield improved performance in terms of predictive accuracy for the identity resolution task across multiple graphs.

- We show that alternative roles exist in social networks (such as in a YouTube graph) and in particular these roles are very different from the known communities in the data.

With respect to multi-relational role discovery we show:

- We propose and study role discovery in multi-relational graphs using tensors and using our novel MRD Tucker decomposition algorithm (see Sections 6 and 7 ).

- We show how to analyze the core tensor of the Tucker decomposition in a multitude of visual and analytic ways to explain the complex interactions occurring (see Section 8).

- We create and measure macro-level properties of the interaction graph such as the simplicity, sharing and stability of the graph with respect to roles (see Table 5).

- We use a constrained formulation of our algorithm that allows *transferring* in knowledge (i.e., roles) from a graph to explain another graph (see Section 9.2) This allows understanding temporal shifts in roles (see Figure 18).

In the next section, we describe related work and then an algorithm for incorporating convex constraints in non-negative matrix decomposition which allows us to encode guidance in a flexible way. Section 4 presents how convex constraints can naturally encode guidance in the form of sparsity and diversity on both the role assignment matrix ($\mathbf{G}$) and role explanation matrix ($\mathbf{F}$). We also present how these constraints can encode the notion of alternativeness to find a different set of roles to another set that are for instance non-actionable or trivial. Our experiments on guidance, in Section 5, demonstrate the usefulness of these forms of guidance in a number of applications and real-world graphs. We show how sparsity and diversity guidance can improve upon prediction performance for the application of identity resolution via roles. We also show how alternativeness can be used to find an alternative set of roles to the underlying community structure. Next in Section 6 we show how multi-relational role discovery can be formulated as a tensor decomposition problem. In particular it can be modeled using a non-negative tucker decompositions, and in that section we also propose our Tucker decomposition algorithm. The Tucker decomposition allows capturing many of the complex interactions between nodes, the roles they play, and the relationships they play them for, which are captured in the core tensor of the Tucker decomposition. In Section 8 we discuss how the core of the Tucker can be interpreted a number of ways, including as a heterogeneous hyper-graph on the space of groups of nodes, groups of features (roles) and groups of relations. Our work opens up many possible novel uses and we experimentally focus on two: i) macroscopic properties of the graphs in terms of roles and ii) transfer settings between multiple graphs which are discussed in Section 9.

## 2 Related Work

The basis for role discovery in graphs using non-negative matrix factorization (NMF) was first proposed in a series of papers at KDD [22][21]. The method ReFeX [22] described a recursive method to take a $n \times n$ adjacency matrix ($A$) and compute a set of $f$ salient features for each of the $n$ nodes represented as a matrix $V$. The RolX method [21] made use of NMF to simplify the features into a set of roles and explored their use for graph matching, sense making and transfer learning. Many previous works had applied NMF to other data mining problems (e.g. [40][28]) but theirs was the first to apply it to role discovery. Other methods for role discovery are not scalable to huge graphs and include Bayesian frameworks using MCMC sampling methods [37] and semi-supervised role labeling [17].

The addition of guidance to matrix decomposition is a relatively new area with most work involving spatial data and properties such as unimodality as we have done for tensors [12]. Of course much work exists on very basic constraints such as non-negativity and minimal rank decompositions. The area of constraints for matrix decomposition takes on several different meanings to our own work. For example in [30] the authors propose the use of labeled information to guide the decomposition. Perhaps the closest to our own work is the use of sparseness constrains in NMF [23].

To the best of our knowledge the encoding of guidance for role discovery and the encoding of diversity and alternative constraints for NMF as described in this paper has not been addressed before. However, the notion of guidance and "alternativeness" is popular in the clustering field with work by ourselves and others [5][36].

## 3  A Constrained NMF Framework for Encoding Guidance into Role Discovery

In this section, we discuss our algorithm for solving the guided role-discovery problem. We present a general algorithm that is well-suited for large-scale problems, and is capable of being extended to different forms of guidance. The different supervisions (described in Section 4) are solvable using this algorithm.

Our algorithm for solving the guided role discovery problem is a *constrained* NMF approach used to find the decomposition shown in Equation 2. Like many unconstrained NMF solvers, it uses the alternating least squares approach [35, 7]. Non-negative least squares is a well-studied problem, and can be utilized to find an NMF solution by solving for one matrix at a time (**G** or **F**), while holding the other constant which is generally known as alternating least squares (ALS). NMF is known to be intractable; and the ALS approach is not guaranteed to find global solutions but will converge to a local minimum. In this work, we add additional constraints to the problem and therefore need more sophisticated methods.

The method we chose was motivated by gradient projection methods, which are known for being well-suited to quickly finding good but not highly accurate solutions for large problems, by sacrificing some of the theoretical convergence guarantees of methods such as interior point [6]. Projected gradient descent methods can be summarized as those that iteratively find better points by following the gradient of the objective function, and subsequently find the closest point that meets the constraints. Since the objective we are solving is least squares, we have a closed form solution to the unconstrained minimum from which we subsequently find the closest constrained solution. It is known, that for a class of constrained least squares solution, this approach will lead to an exact global solution in one iteration (see Lemma 1).

Therefore, our algorithm has the advantage that each *subproblem* (but not the entire problem) can be solved exactly by reducing it into an unconstrained least square problem [39][3] and an Euclidean projection problem [14][32], both of which have efficient solutions. Additionally, this approach to optimization (projected gradient descent) has been shown in the past to work well on large-scale problems, at the expense of accuracy, and is used by state of the art solvers [31].

The outline of the remainder of this section is as follows. First, we formally describe the convex constrained NMF problem and discuss how ALS can be used to solve it. Then, we explain how ALS can also be used to solve for individual role assignment vectors, as well as role definition vectors. Finally, we describe how ALS over definition/assignment vectors can be solved using a projection method by first solving an unconstrained least squares problem and then finding the closest point in the constrained space.

**The Constrained NMF Problem** In Equation 2, there are two variables **G** and **F** that are being simultaneously optimized. If either is treated as a constant, the problem becomes convex and can be solved exactly using any method for solving convex optimization problems. One can alternate between solving for **G** and **F** this way until convergence. Although each iteration finds a global optimum to this modified problem, the result of this procedure (alternating optimization) is not guaranteed to find a global minimum to the original problem in Equation 2. In the following, we describe our method for transforming the formulation into a series of convex programming problems, which are generally easy to solve.

$$\begin{aligned} \underset{\mathbf{G},\mathbf{F}}{\text{minimize}} \quad & ||\mathbf{V}-\mathbf{GF}||_2 \\ \text{subject to} \quad & g_i(\mathbf{G}) \leq d_{Gi},\ i=1,\ldots,t_G \\ & f_i(\mathbf{F}) \leq d_{Fi},\ i=1,\ldots,t_F \end{aligned} \quad (2)$$

where $g_i$ and $f_i$ are convex functions.

**An ALS Formulation** Rather than alternating between solving for the entire matrices **G** and **F**, we can instead solve for one column of **G** or one row of **F** at a time. This is possible if convex constraints can be specified in terms of these columns, which is the case in this work. Without loss of generality, Equation 3 shows an individual sub-optimization problem in terms of one of the columns of **G**, denoted **x**.

$$\mathbf{G}_k = \underset{\mathbf{x}}{\text{minimize}} \quad ||\mathbf{R}-\mathbf{xF}_k||_2 \\ \text{subject to:} \quad g_i(\mathbf{x}) \leq d_{Gi},\ i=1,\ldots,t_G \quad (3)$$

In Equation 3, **R** represents the residuals of all other factors not being solved for (sum of outer products of corresponding columns of **G** and rows **F**). $\mathbf{F}_k$ is the $k^{th}$ row of the role/feature explanation matrix that corresponds to the $k^{th}$ column of the role assignment matrix. So with this formulation, we alternate between learning single role assignments, followed by learning a role definition. Next we explain how we solve the convex constrained problem shown in Equation 3.

**Solving The Constrained Least Squares Problem** Our projection method is as follows. First, solve Equation 3 with all constraints removed using standard least squares solvers. Second, find the closest point to the unconstrained solution, that satisfies the given constraints. This projection method takes advantage of standard and very fast least squares solvers and the subsequent nearest feasible point problem is relatively simple to solve. In addition, Lemma 1 shows that performing these two steps will exactly solve the original prob-lem in Equation 3. Applications of this theorem and its proof can be found in [10][20].

**Lemma 1** *Projection Equivalence Result. The following constrained optimization problem:*

$$\begin{aligned} \underset{\mathbf{x}}{minimize} \quad & ||\mathbf{B} - \mathbf{xa}||_2 \\ subject\ to: \quad & c_i(\mathbf{x}) \leq d_i,\ i = 1, \ldots, n \end{aligned} \quad (4)$$

*where $c_i$ are convex functions on $x$, is equivalent to:*

$$\begin{aligned} \underset{\mathbf{x}}{minimize} \quad & ||\mathbf{x}^* - \mathbf{x}||_2 \\ subject\ to: \quad & c_i(\mathbf{x}) \leq d_i,\ i = 1, \ldots, n \end{aligned} \quad (5)$$

*where $\mathbf{x}^*$ is the optimal to the optimization problem in Equation 4 without contraints.*

This leads to the following algorithm for convex constrained NMF presented in Figure 1. Like ALS for unconstrained NMF, this heuristic is not guaranteed to meet a global optimum, even though all subproblems are solved exactly. However, each step will lead to a reduction in the global objective (Equation 2). Thus, in practice the algorithm will find local minima that meet all specified constraints.

The advantage of solving for one role at a time rather than the entirety of **G** or **F** as is generally done with ALS, is that it allows the problem to be broken down into smaller parts that then fit into fast solvers. In general, projection methods have been found to be better suited to larger problems and we found this to be the case as well. Using this method allows us to solve much larger problems than we had previously been able to using standard constrained optimization solvers [12]. The final constrained optimization problem (i.e., closest constrained point problem) is simple enough that we find for even medium-sized problems we could utilize high level solvers such as CVX [11][19], which makes experimenting with new types of constraints very simple.

## 4 Framework for Flexible Supervision

In the previous section, we discussed a novel and general algorithm that can easily handle convex constraints. Convex constraints can encode a variety of useful guidances. In this section, we show how they can be used to enforce sparsity, diversity and alternativeness. In the experimental section, we show applications which exploit these forms of guidance.

### 4.1 Sparsity

The area of sparsity has recently attracted much attention. In a general context, sparsity has been shown to have two main benefits: (1) parsimony and (2) improved predictive performance, with the later being motivated by Occam's

Table 1: Summary of effects of constraints on both role assignment **G** and role definitions **F** (see Section 4 for formulation of each constraint type).

|  | **Role Assignment** | **Role Definition** |
|---|---|---|
| **Sparsity** | Encourages role assignments to be more definitive. Increasing the strength of constraint reduces the number of nodes that have minority membership in role. | Increases the ability to interpret role definitions by ensuring that the definitions only use features most strongly correlated with each role. Increasing the strength of constraint decreases the likelihood that features with small explanatory benefit are included. |
| **Diversity** | Roles cannot have memberships that are too similar. No two roles can have exactly the same membership assignment. Increasing the strength of the constraint limits the amount of allowable overlap in assignments. | Roles cannot have definitions that are too similar. No two roles can have redundant explanations and increasing the strength of constraint ensures that roles must be explained with completely different sets of features. |
| **Alternative** | Find a set of roles that lends itself to a different role assignment than a given role assignment. Increasing the strength of constraint, decreases the allowable similarity between the two. | Learn a role definition matrix that is significantly different than a given role definition. Increasing the strength of constraint ensures that the definitions must be very dissimilar. |

**Inputs:**

- **V**: Node feature matrix containing $n$ nodes described by $f$ topological structure features.
- $g_i(\mathbf{x}), f_i(\mathbf{x})$: Convex constraints on columns of **G** and rows of **F** respectively.
- $r$: Number of roles (methods for learning $r$ described in previous work [21]).

**Outputs:**

- **G**: Role assignment matrix that satisfying all constraints.
- **F**: Role definition matrix that satisfying all constraints.

**Algorithm:**

**while** *reconstruction error decreases* **do**
{
   **for** $k = 1 \ldots r$   //*Recalculate each role.*
   {

1. Calculate $\mathbf{R} = \mathbf{V} - \mathbf{G}_{\bullet(\neq k)}\mathbf{F}_{(\neq k)\bullet}$

2. Calculate $\mathbf{G}_{\bullet k}$ by solving for **x** as follows:
   - (a) $\mathbf{x}^* = \underset{\mathbf{x}}{\operatorname{argmin}} \; ||\mathbf{R} - \mathbf{x}\mathbf{F}_{\mathbf{k}\bullet}||_2$
   - (b) $\mathbf{G}_{\bullet k} = \underset{\mathbf{x}}{\operatorname{argmin}} \; ||\mathbf{x}^* - \mathbf{x}||_2$ s.t. $g_i(\mathbf{x}) \leq \epsilon_i : \forall_i$

3. Calculate $\mathbf{F}_{k\bullet}$ by solving for **x** as follows:
   - (a) $\mathbf{x}^* = \underset{\mathbf{x}}{\operatorname{argmin}} \; ||\mathbf{R} - \mathbf{x}\mathbf{G}_{\bullet \mathbf{k}}||_2$
   - (b) $\mathbf{F}_{k\bullet} = \underset{\mathbf{x}}{\operatorname{argmin}} \; ||\mathbf{x}^* - \mathbf{x}||_2$ s.t. $f_i(\mathbf{x}) \leq \epsilon_i : \forall_i$

   }
}

Figure 1: Our algorithm that will be used to encode all guidances described in Section 4. The algorithm uses a least squares approach and allows additional convex constraints to be added to the NMF formulation.

razor. Sparse learning formulations exist for many learning settings such as linear regression (LASSO), Kernel methods (SVM) and covariance estimation.

In our work, we can place sparsity constraints on both the **G** or **F** matrices leading to an objective function of:

$$
\begin{aligned}
\operatorname*{argmin}_{\mathbf{G},\mathbf{F}} \quad & ||\mathbf{V} - \mathbf{GF}||_2 \\
\text{subject to:} \quad & \mathbf{G} \geq 0, \mathbf{F} \geq 0 \\
& \forall i \quad ||\mathbf{G}_{\bullet \mathbf{i}}||_1 \leq \epsilon_G \\
& \forall i \quad ||\mathbf{F}_{\mathbf{i} \bullet}||_1 \leq \epsilon_F \\
\text{where} \quad & \epsilon_G \text{ and } \epsilon_F \text{ define upperbounds for the sparsity} \\
& \text{constraints (amount of allowable density).}
\end{aligned}
\tag{6}
$$

Previous works have shown the effectiveness of using L1 norm as a penalty in model learning. In our formulation the L1 penalty is encoded as a constraint rather than a penalty in the objective, but it is known that these formulations are theoretically equivalent [8]. However, another twist to our formulation is that we do not constrain the entire matrix but instead constrain each column of **G** and each row of **F**. This was done because our solver requires constraints to be formulated only over one role vector at a time. The effect of this technical difference is that the sparsity must be more uniformly spread across each role definition or role assignment which is a benefit of this method.

Sparsity constraints on **G** and **F** have easy to understand intuitive interpretations. If **G** is sparse, it means that nodes are assigned to as few roles as possible; and it is possible for some nodes to be assigned to no roles. If **F** is sparse, it means that the roles are defined with respect to as few features as possible. Both of these extensions allow for a simple explanation of the data, and lead to improved prediction performance.

### 4.2 Diversity

In the NMF forms of role discovery, nothing prevents the roles to which nodes are assigned (i.e., the **G** matrix) and the role definitions (i.e., the **F** matrix) to be highly overlapping. This can be undesirable particularly for the **F** matrix since it means all roles are highly similar. This can be overcome by enforcing a diversity requirement so that each role uses a different set of features (for the **F** matrix) and nodes are assigned to different combinations of roles (for the **G** matrix).

Our formulation for role allocation diversity (**G** matrix) and role definition diversity (**F** matrix) makes use of orthogonality as follows:

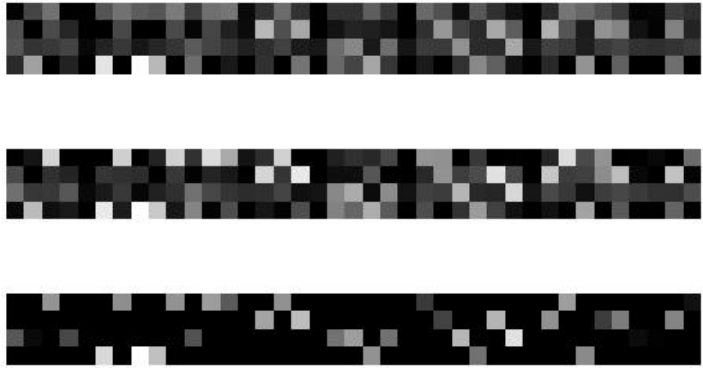

Figure 2: Visualization of diversity constraints on role explanation matrix $F$ (roles × features) for DBLP dataset. The top matrix shows the unconstrained result; the bottom matrix is constrained to be completely diverse ($\epsilon = 0$); and the middle matrix shows a middle ground. From the top matrix to the bottom matrix, the number of black cells (i.e. zero values) increases since roles definitions must be explained with completely different sets of features.

$$
\begin{aligned}
\operatorname*{argmin}_{\mathbf{G},\mathbf{F}} \quad & ||\mathbf{V} - \mathbf{GF}||_2 \\
\text{subject to:} \quad & \mathbf{G} \geq 0, \mathbf{F} \geq 0 \\
& \forall i,j \quad \mathbf{G}_{\bullet i}^T \mathbf{G}_{\bullet j} \leq \epsilon_G \quad i \neq j \\
& \forall i,j \quad \mathbf{F}_{i\bullet} \mathbf{F}_{j\bullet}^T \leq \epsilon_F \quad i \neq j \\
\text{where} \quad & \epsilon_G \text{ and } \epsilon_F \text{ define upperbounds on how angularly} \\
& \text{similar role assignments and role definitions can} \\
& \text{be to each other.}
\end{aligned}
\quad (7)
$$

When $\epsilon = 0$, our constraint will exactly match the definition orthogality, and when $\epsilon \geq 0$ the constraint can be viewed as limiting the angular similarity between two vectors. The effect of combining this constraint with non-negativity constraints is that no role definitions will have any common features and no role assignments will have overlapping populations for $\epsilon = 0$. This is so since $\mathbf{G}_{\bullet i}^T \mathbf{G}_{\bullet j} = 0$ if and only if these two vectors do not share any non-zero entries. Figure 2 shows such an example, where none of the three roles have any overlapping features. In the context of our solver which solves for one vector at a time, this constraint will be linear (a weighted sum).

## 4.3 Alternative Role Discovery

Recent work on another unsupervised problem, clustering, has explored the area of alternativeness [36, 13]. In that literature, the term alternativeness and orthogonality are used interchangeably, but we only use the term alternativeness for clarity.

The motivation for alternativeness in unsupervised learning is strong. Most interesting problems are on large data sets that contain complex phenomena and there may exist multiple explanations of the data. However, most unsupervised learning algorithms expect that there exists only one good explanation of the data and return one explanation.

In many situations, it may be the case that the returned explanation is undesirable since it is either unactionable or not novel. Consider the IMDB (Internet Movies Database) dataset. If the resultant roles map actors to the studios for which they work, then this is not particularly novel. Here, the work on alternative role discovery allows a previously discovered set of role allocations ($\mathbf{G}^*$) or role definitions ($\mathbf{F}^*$) to be specified as a counter-example of what not to find. The challenge though is to find another good explanation of the data that is different to those already found.

The optimization problem to find alternative roles is then:

$$
\begin{aligned}
& \underset{\mathbf{G}, \mathbf{F}}{\operatorname{argmin}} && ||\mathbf{V} - \mathbf{GF}||_2 \\
& \text{subject to:} && \mathbf{G} \geq 0, \mathbf{F} \geq 0 \\
& && \forall i, j \quad \mathbf{G}^{*T}_{\bullet i} \mathbf{G}_{\bullet j} \leq \epsilon_G \\
& && \forall i, j \quad \mathbf{F}^*_{i \bullet} \mathbf{F}^T_{j \bullet} \leq \epsilon_F \\
& \text{where} && \epsilon_G \text{ and } \epsilon_F \text{ define upperbounds on how similar} \\
& && \text{the results can be to } \mathbf{G}^* \text{ and } \mathbf{F}^*.
\end{aligned}
\tag{8}
$$

# 5 Experiments for Guided Role Discovery

Our experiments demonstrate how constraints on graph role discovery can be useful. Role discovery requires the user to specify the number of roles to use and a set of features for a graph. For the former, we used the Minimum Description Length (MDL) described in [21] to automatically select the number of roles; and for the later, we used the approach described in [22]. We show that role discovery can be used to improve the results of the identity resolution problem between two graphs, and that they can be further improved by using sparsity or diversity constraints. By using sparsity or diversity constraints, we improve the role definitions which leads to more meaningful role assignments and more accurate identity resolutions. See Section 5.1 for these experiments. We also experimentally verify the solutions to the alternative role discovery formulation presented in Section 4.3 and observe that they indeed produce significantly different results. The purpose of our experimental section is to address the following questions:

| Network | —V— | —E— | k | —LCC— | #CC |
|---------|-----|-----|------|-------|-----|
| **VLDB**   | 1,306 | 3,224 | 4.94 | 769   | 112 |
| **SIGMOD** | 1,545 | 4,191 | 5.43 | 1,092 | 116 |
| **CIKM**   | 2,367 | 4,388 | 3.71 | 890   | 361 |
| **SIGKDD** | 1,529 | 3,158 | 4.13 | 743   | 189 |
| **ICDM**   | 1,651 | 2,883 | 3.49 | 458   | 281 |
| **SDM**    | 915   | 1,501 | 3.28 | 243   | 165 |

Table 2: Information about DBLP co-author networks for each conference. Data was collected for five years (2005-2009). —V—=number of vertices, —E—=number of edges, k=average degree, —LCC—=size of largest connected component, #CC=number of connected components.

1. Does adding constraints to the NMF-based role discovery formulation improve the quality of the resulting role explanations and assignments? Figures 3 and 4 show that constraints improve the results of identity resolution.

2. What effects do diversity constraints have on role discovery results? Figures 3 and 4 show how diversity constraints can improve role discovery results even more so than sparsity constraints.

3. Can our alternative role discovery formulation produce significantly different results? Tables 3 and 4 shows that our formulation can produce results that are significantly different than a given set of roles or community assignments respectively.

## 5.1 Sparse and Diverse Identity Resolution in Co-authorship Graphs

In this experiment, we show that by adding sparsity and diversity constraints to the NMF formulation of role discovery, the resulting role definitions are of higher quality. We measure this improvement in quality indirectly by showing how role definition matrices can be used for resolving identities of nodes across graphs, and that constrained role definitions perform better than unconstrained role definitions for that problem.

From the DBLP data-set [27], we extracted a co-author graph from each of the following related conferences from 2005 to 2009: KDD, ICDM, SDM, CIKM, SIGMOD, VLDB (see Table 2 for detailed information about each co-author graph). We extract a set of relevant structure features for the KDD graph using REFEX [22], and compute these same features for all of the co-author graphs. We subsequently learn a set of role definitions from the KDD graph using standard RolX [21] as well as the sparse and diverse versions of GLRD. For each of these competing role definitions, we assign each vertex from each graph to the roles whose function they most exhibit. As a baseline, we also explore author identification without roles by using the raw graph features as described in ReFeX.

We use the role assignments to resolve the identities of vertices from each graph (namely, ICDM, SDM, CIKM, SIGMOD, and VLDB) to the vertices in the KDD graph. Without loss of generality, assume we are resolving identity of authors from the KDD graph to the authors in ICDM graph. For each author in both conferences, we select the corresponding row vector from the node by role matrix $\mathbf{G}_{kdd}$ and find the $k$ closest neighbors (row vectors) from $\mathbf{G}_{icdm}$. If the original author from KDD graph is present in the set of $k$ closest neighbors, we count the result as a match. We repeat this experiment using sparsity and diversity constraints on $\mathbf{F}_{kdd}$. We also repeat the experiment using the ReFeX features, comparing author feature vectors from $\mathbf{V}_{kdd}$ and $\mathbf{V}_{icdm}$. Figures 3 and 4 shows how the different decomposition methods compare in this setting for all graphs paired with KDD.

Our method of utilizing role discovery results for the author identification task is described formally in the following set of steps:

1. Extract features from co-authorship graphs to get graph features (e.g. $\mathbf{V}_{kdd}, \mathbf{V}_{icdm}$) using ReFeX.

2. From the graph features matrix $\mathbf{V}_{kdd}$ perform role discovery to obtain $\mathbf{G}_{kdd}$ and $\mathbf{F}_{kdd}$.

3. Transfer the role definition matrix $\mathbf{F}_{kdd}$ (role by feature matrix) to other graphs (e.g. $\mathbf{V}_{icdm}$) by solving Equation 9.

$$\mathbf{G}_{icdm} = \min_{\mathbf{G}} ||\mathbf{V}_{icdm} - \mathbf{G}\mathbf{F}_{kdd}||_2 \text{ s.t. } \mathbf{G} \geq 0 \tag{9}$$

Our experiments with graph identity-resolution show that diversity and sparseness constraints almost universally improve the quality of learned role-definition matrix. This is not unexpected since there is a long tradition in machine learning of using sparsity to prevent overfitting. As mentioned previously we can view diversity as enforcing sparsity since a diverse set of roles as per our definition do not share many overlapping features and hence each role definition is concise.

Figure 3 shows that role definitions learned using sparsity and diversity outperform standard unconstrained role discovery (RolX) in almost every setting and problem parameterization. Figure 4 more clearly shows the general trend by considering the results for a particular problem parameterization. In that figure, we observe that diversity constraints lead to the most improvement over RolX, while sparsity improvements are lesser. We also observe that transferring the KDD role definitions to some graphs (like VLDB and SIGMOD) does not compare well to the baseline method that does not use any roles (such as ReFeX). We believe this is because the same participants in conferences such as VLDB and SIGMOD do not have a similar role to the ones they play in KDD; and hence, using the raw features (without roles) produces better results.

We believe that sparsity improves the quality of role definitions by reducing the ability of unconstrained NMF-based role discovery to overfit the problem. Features that only slightly add to the definition of a role are more likely to be explaining noise; and by forcing those values to zero, we end up with more robust definitions. Furthermore, the diversity constraints help by removing redundancy in role definitions, which leads to definitions that are more easily comparable. For example, if a feature is used to define every role, then it is not essential in defining any of them.

## 5.2 Alternative Roles

In this section, we show that our alternative role discovery formulation (presented in Section 4.3) can discover significantly different role definitions, as well as show that the formulation can be used to improve the role definitions when there are ground-truth communities. In Table 3, we show the difference between an alternative role discovery result and an original role definition found using unconstrained role discovery (via RolX). In Table 4, we show that we can use our formulation to get more consistent assignments of roles when ground-truth communities are known.

In our first experiment, we explore the difference between the roles of the original and alternative role discovery. Using the KDD co-authorship graph, we find a set of roles and constrain a new solution to have a significantly different role definition ($\mathbf{F}$ matrix). We then compare the results by assigning each vertex to its most dominant role in both results to create two separate partitions of the vertices. We then measure the difference between the two partitions using Jaccard distance. Table 3 shows that all of the Jaccard distances are far from 0 meaning that the alternative role assignments are very different than the original ones. Figure 5 illustrates the alternative roles found in the largest connected component of the KDD coauthorship graph. Note, the reader can zoom in on this figure to read the names of each author. The following is a description of the original roles and the roles that GLRD(Alternative) found. These description are based on sense-making analysis [21]. As the descriptions show these roles are capturing alternative concepts.

|    | R1(alt) | R2(alt) | R3(alt) | R4(alt) |
|----|---------|---------|---------|---------|
| R1 | 0.946   | 0.510   | 0.762   | 0.913   |
| R2 | 1.000   | 0.971   | 0.810   | 0.739   |
| R3 | 1.000   | 0.7942  | 1.000   | 1.000   |
| R4 | 0.345   | 0.991   | 1.000   | 0.982   |

Table 3: Jaccard distance matrix comparing original role assignments (rows) to alternative role assignments (columns). Jaccard distance of 0 represents an exact match between clustering and 1 represents no overlap. The relative error for the two decompositions was similar: 0.12% and .5% (where relative error is $error = ||\mathbf{V} - \mathbf{GF}||/||\mathbf{V}||$).

**Original Roles:**

- **Role 1:** Nodes here have high eccentricity. These are periphery nodes.

- **Role 2:** Nodes here have high eccentricity and high clustering coefficient. These are periphery nodes that are cliquey.

- **Role 3:** Nodes here have high degree and high clustering coefficient. These are highly connected cliquey nodes.

- **Role 4:** Nodes here have high PageRank, high degree, and high biconnected components numbers. These are globally central stars and brokers.

**Alternative Roles:**

- **Role 1:** Nodes here have high PageRank and high biconnected component numbers. These are globally central and brokers.

- **Role 2:** Nodes here have high clustering coefficient but not high eccentricity. These are non-periphery nodes that are cliquey.

- **Role 3:** Nodes here have high eccentricity and high clustering coefficient. These are periphery nodes that are cliquey.

- **Role 4:** Nodes here have high eccentricity and high degree. These are periphery nodes that are locally stars.

We next experiment with a YouTube dataset, which is a network of users with known ground-truth communities [33]. This graph was created by crawling the YouTube site in 2007 and creating directed edges between a pair of users $a$ and $b$ if $a$'s profile page linked to $b$'s profile page. Ground-truth communities were assigned by collecting all users belonging to the same group, which were pages that allowed communications between users on given topics. The graph has 1,134,890 vertices, 2,987,624 edges, and 8,385 communities. We selected all communities with over 100 users of which there were 105. The largest community has 2,217 users.

There is an inherent complementariness between role discovery and community detection. The former is about structural similarity; while the latter is based on proximity in the graph. Role discovery finds functions/roles of users but does not find the communities themselves. However, there may be multiple interesting sets of communities within the same network and those communities may be characterized by very different roles. In this experiment, we encode the set of ground-truth communities for which our role discovery technique should find roles.

The way we encode the YouTube ground-truth communities into our analysis is by providing the communities as $\mathbf{G}^*$ to our alternative role discovery formulation. This will force our discovered roles to have a role assignment that is different than ground-truth communities, which matches the semantic relationship between the two problems.

To evaluate the effectiveness of this result we measured the proportion of members in each community belonging to each role. We then calculated the standard deviation over all such communities per role and report the results in Table 4. The assumption for this evaluation is that each role should be equally represented in each community. Our results show that the alternative role discovery formulation can indeed be used to normalize the roles with respect to a set of ground-truth communities. After applying sense-making [21], the six roles that our GLRD(Alternative) finds are as follows:

**Alternative Role 1:** Nodes here are global hubs. They have high PageRank values, high out-degrees, and high biconnected component numbers.

**Alternative Role 2:** Nodes here are on the periphery of the graph. They have higher than default eccentricity.

**Alternative Role 3:** Nodes here are authorities. They have high PageRank values and high in-degrees.

**Alternative Role 4:** Nodes here are very cliquey. They have high clustering coefficients.

**Alternative Role 5:** Nodes here are local hubs. They have high out-degrees and high biconnected component numbers.

**Alternative Role 6:** Nodes here are the majority of the population; they are the "regular" folks. They have a local neighborhood that is more cliquey than expected but otherwise nothing special stands out.

| Roles | 1 | 2 | 3 | 4 | 5 | 6 |
|---|---|---|---|---|---|---|
| Original | 7.85 | 7.93 | 8.70 | 2.35 | 9.81 | 7.57 |
| Alternate | 5.06 | 6.34 | 5.34 | 3.81 | 8.62 | 5.88 |

Table 4: For each role, we report the standard deviations of role proportions over all communities. The result shows that our alternative role discovery formulation can be used to find roles whose members are better distributed across a set of interesting communities. The values are scaled by $10^2$.

# 6 Lifting our Formulation for Multi-Relationl Role Discovery

Here we outline our method to lift our previous work to perform role discovery in multi-relational graphs. We do not recreate the same experiments since they are trival but instead focus on the more challenging problem of role discovery in multi-relational graphs.

**Role Discovery in Multi-relational Graphs.** Our approach to extending role discovery to multi-relational graph is to model the graphs as a tensor. This is done by extracting features from each relation and appending the resulting feature matrices into a single tensor $\mathcal{V}$ of dimension $n \times f \times r$. Just as NMF is used to decompose a feature matrix $V$, tensor decompositions can be used to decompose a feature tensor $\mathcal{V}$. One natural choice of tensor decompositions to decompose a feature tensor would be non-negative PARAFAC [16]. PARAFAC like NMF is a rank one decomposition see Figure 6. However, PARAFAC is not an ideal model to find complex patterns in graphs, as is desired for role discovery, because it is too simplistic in its assumptions. In particular it will only allow each group of entities to play only one role for only one group of relations. See the introductory section for a more indepth explanation of the limitations of PARAFAC.

$$\operatorname*{argmin}_{\mathbf{G},\mathbf{F},\mathbf{R}} \quad ||\mathcal{V} - \sum_k \mathbf{g_k} \circ \mathbf{f_k} \circ \mathbf{r_k}||_{Fro}$$
$$\text{subject to:} \quad \mathbf{G} \geq \mathbf{0}, \mathbf{F} \geq \mathbf{0}, \mathbf{R} \geq \mathbf{0} \tag{10}$$

Instead we use the Tucker decomposition (shown in Equation 11) that allows us to find the complex interaction between E-groups, the roles they play, and R-groups they play those roles in. The diagrammatic explanation of Tucker decomposition in Figure 7 shows how it models these interactions. Like PARAFAC and NMF, it is a rank one decomposition which allows for an intuitive interpretation. A column in $G$ corresponds to a group of people and is a length $n$ indicator vector showing E-group membership. Similarly a column in $F$ corresponds to a role definition which is a group of features and a column in $R$ corresponds to a group of relations which we refer to as an R-group. Unlike PARAFAC and NMF, any factor can be any combination of the columns in $G$, $F$, and $R$. The core of the Tucker decomposition allows this complex interaction and requires more explanation (PARAFAC can be viewed as a specific Tucker with diagonal core). It too is a order 3 tensor except the modes are now directly interpretable as E-groups, roles, and R-groups. An entry in the core at $i, j, k$ means that E-group $i$ plays role $j$ for R-group $k$. Understanding and simplifying this core is critical to the success of multi-relational role discovery using a Tucker decomposition.

$$\operatorname*{argmin}_{\mathbf{G},\mathbf{F},\mathbf{R},\mathcal{H}} \quad ||\mathcal{V} - \sum_i \sum_j \sum_k h_{ijk} * \mathbf{g_k} \circ \mathbf{f_k} \circ \mathbf{r_k}||_{Fro}$$
$$\text{subject to:} \quad \mathbf{G} \geq \mathbf{0}, \mathbf{F} \geq \mathbf{0}, \mathbf{R} \geq \mathbf{0}, \mathcal{H} \geq \mathbf{0} \tag{11}$$

# 7 Our MRD Algorithm For Multi-Relational Graphs

The Tucker model has most often been described as a higher order analog of principal component analysis or singular value decomposition and is traditionally defined with factor matrices being orthogonal. Among the most popular tensor toolboxes, the Tucker model is often implemented with orthogonality constraint on the factor matrices (Tensor Toolbox [4, 2]) or with no constraint enforced on the core (Nway Toolbox [1]). Other recently proposed algorithms for non-negative Tucker model [24, 34] extend the classical multiplicative update procedures proposed for NMF [26], which is known to converge slowly near stationary points [29]. Since the alternating least squares (ALS) method is known as the "workhorse" algorithm for PARAFAC [25] and is empirically demonstrated to be competitive among many existing methods [38], we implement our own version of non-negative Tucker decomposition using an alternating non-negative least squares (ANLS) scheme.

Let $\mathcal{V}$ be the tensor to be decomposed. Denote the factor matrices by $G, F$ and $R$ and the core tensor by $\mathcal{H}$. In each iteration we optimize over each of $G, F, R$ and $\mathcal{H}$ in turn while fixing all others as constants. When $G$ is being optimized, the objective can be written as:

$$\underset{\mathbf{G} \geq \mathbf{0}}{\operatorname{argmin}} \quad \|\mathcal{V}_G - \mathbf{G}\mathcal{H}_G(\mathbf{R} \otimes \mathbf{F})^T\|_{Fro} \tag{12}$$

where $\mathcal{V}_G$ is the matricization of $\mathcal{V}$ in the first mode and $\otimes$ is the Kronecker product. The subproblems when $F$ and $R$ are being solved for have the exact same form but with a different variable being optimized. In addition it is generally desirable for the entries in the core to indicate the weights of each coupling of factors. Thus we normalize the columns of $G, F$ and $R$ once they are solved. When we solve for the core $\mathcal{H}$, rewriting the tensors in vectorized form turns the objective into:

$$\underset{\mathcal{H} \geq \mathbf{0}}{\operatorname{argmin}} \quad \|\text{vec}(\mathcal{V}) - (\mathbf{R} \otimes \mathbf{F} \otimes \mathbf{G})\text{vec}(\mathcal{H})\|_{Fro} \tag{13}$$

where $\text{vec}(\cdot)$ is the vectorization of a tensor. Our overall solver is summarized in Algorithm 1. We build our solver on top of the existing constructs in the MATLAB tensor toolbox [2] and employ the fast non-negative least squares (NNLS) solver particularly designed for tensor decomposition [9] when we solve subproblems (12) and (13). For the terminating condition we adopt the common practice for ALS which stops when the relative change in the objective between successive iterations is smaller than some pre-set threshold. It is worth noting that although we only enforce non-negativity constraints in this case, it requires little effort to adopt any constraint applicable to standard least squares problem into our formulation.

**Algorithm 1** Multi-relational Role Discovery (MRD) using Alternating Least Squares Non-negative Tucker decomposition.

1: Initialize $\mathbf{G}, \mathbf{F}, \mathbf{R}$ and $\mathcal{H}$ to any non-negative values
2: **while** Stop condition not met **do**
3:     $\mathbf{G} \leftarrow \underset{\mathbf{G} \geq \mathbf{0}}{\operatorname{argmin}} \quad \|\mathcal{V}_G - \mathbf{G}\mathcal{H}_G(\mathbf{R} \otimes \mathbf{F})^T\|_{Fro}$
4:     Normalize the columns of $\mathbf{G}$
5:     $\mathbf{F} \leftarrow \underset{\mathbf{F} \geq \mathbf{0}}{\operatorname{argmin}} \quad \|\mathcal{V}_F - \mathbf{F}\mathcal{H}_F(\mathbf{R} \otimes \mathbf{G})^T\|_{Fro}$
6:     Normalize the columns of $\mathbf{F}$
7:     $\mathbf{R} \leftarrow \underset{\mathbf{R} \geq \mathbf{0}}{\operatorname{argmin}} \quad \|\mathcal{V}_R - \mathbf{R}\mathcal{H}_R(\mathbf{F} \otimes \mathbf{G})^T\|_{Fro}$
8:     Normalize the columns of $\mathbf{R}$
9:     $\mathcal{H} \leftarrow \underset{\mathcal{H} \geq \mathbf{0}}{\operatorname{argmin}} \quad \|\operatorname{vec}(\mathcal{V}) - (\mathbf{R} \otimes \mathbf{F} \otimes \mathbf{G})\operatorname{vec}(\mathcal{H})\|_{Fro}$
10: **end while**
11: **return** $\mathbf{G}, \mathbf{F}, \mathbf{R}, \mathcal{H}$

**Algorithm Complexity.** Our algorithm is an example of alternating least squares with each step being efficiently solvable using least squares solvers. The non-negativity requirement on the core can be efficiently enforced by solvers. Since tensor decomposition is well known to be intractable, we provide an estimate of our algorithm's run time to converge to a good local minima. The algorithm like most tensor decomposition algorithms has linear complexity with respect to the number of factors, modes and size of the core. In practice the decomposition of our graphs shown in the experimental section took under a minute to run on a 12-core machine.

## 8 Interpretting Tensor Decomposition for Role Discovery

After applying Algorithm 1 we have decomposed the multi-relational graph into a series of E-groups (defined by $G$), a series of roles (defined by $F$) and a series of R-groups (defined by $R$). The core of the Tucker decomposition measures the interaction between these E-groups, roles and R-groups. Here we show how to interpret and analyze the results of Tucker decomposition in a number of ways.

### 8.1 Visually Interpreting Core Slices

We begin with the simple but useful approach of visually inspecting the core tensor slices to compare E-groups, roles, or R-groups. A slice of the core (depending on its orientation: left-to-right, top-to-down or back-to-front) can represent a E-group, role, or R-group. Different slices of the same orientation can then be used to compare the similarity of E-groups, roles and R-groups. For example in Figure 8 we display the slices corresponding to different E-groups from a multi-relational role discovery result.

Comparing the slices directly leads to very detailed comparison of E-groups because we compare for example if they have role/R-group combinations in common. However if we consider aggregations of these slices we can get more coarse comparison, such as whether or not the E-groups play the same roles, or whether they participate in the same R-groups. For example the third and fifth E-group look very similar in terms of the R-groups they take part in, but by looking at the slices we know that they differ because they play very different roles in those very same relations.

## 8.2 Visualizing Core as an Interaction Graph

A further visual understanding of the phenomenon in the multi-relational graph can be obtained by visualizing the core as a graph. This is achieved by creating a node for every E-group, role, and R-group. This will of course be a heterogeneous graph. An entry in the core then could be represented in this graph as a clique on the triplet (E-group, role, R-group) it corresponds to. Since each edge corresponds to a Tucker core entry, it's edge can be weighted depending on that core value entry and be interpreted as a similarity. However, if we are focused say on predominantly understanding groups of entities, we can create a tripartite graph as shown in Figure 9 which removes the edge between the role and R-group. We shall call this graph the **interaction graph** to distinguish it from the original multi-relational graph we study.

This interaction graph can then be visualized and interesting signature patterns can be interpreted. See Figure 9 for some example signatures.

## 8.3 Analysis of the Interaction Graph

Given the interaction graph described in the previous subsection which shows the relationship between E-groups, roles, and R-groups, we can analyze this graph any number of ways. For example, a popular approach to graph simplification is to embed the graph into a two dimensional space. Figure 16 shows such an embedding using PCA of the graph written in "hyper-edge" form. That is a $n \times m$ matrix where each column in the matrix represents a hyper-edge and entry $i, j$ has value 1 if node $i$ is involved in hyper-edge $j$. This heterogeneous object embedding can be interpreted such that each cluster is a collection of E-groups, roles, and R-groups that often interact.

| Property | Computation |
|---|---|
| Simplicity: To what extent are nodes connected to multiple nodes of other types versus being connected to only one node (e.g., E-groups playing multiple roles)? | Average node degree |
| Sharing: How much can E-groups be separated into independent parts? For example, can we find two sets of roles that are played by completely non-overlapping sets of E-groups? | Mincut cost |
| Variability: How does the simplicity of nodes (E-groups, roles, or R-groups) vary across the interaction graph. | Variance of node degree; Entropy of PageRank distribution |
| Stability: How stable are the interactions between roles, E-groups and R-groups | Spectral gap |

Table 5: The macroscopic properties measured on the interaction graph $\mathcal{H}$. See Figure 17 for measurements over several congressional multi-relational graphs spanning a time frame of 30 years.

## 8.4 Macroscopic Properties Derived from the Interaction Graph

Given the interpretation of the core as an interaction graph, we can than understand the macroscopic properties of the role dynamics by analyzing the interaction graph properties. The metrics we study are motivated in Table 5 along with how they are computed. These metric are meant to give the user a broad understanding of the underlying dynamics of the graph. The simplicity property tells how strongly aligned E-groups, roles and R-groups are, while the sharing property measure how many roles, and R-groups, are shared among different groups of entities. The variability property, captures the amount of imbalance in the complexities of different nodes in the interaction graph, by calculating both the variance of the node degrees as well as the entropy of the stationary distribution on a random walk along the interaction graph. Another important property we measure is the stability of the results we discovered. Here we wish to answer the question, how robust are the patterns found within the interactions graph and how easily could those patterns change due to small perturbations.

## 8.5 Complex Analysis Via Role Transfer

Our work so far learned both the E-groups, role definitions and R-groups from the **one** multi-relational graph. However, we can transfer in these definitions from another source by holding them fixed as constants in the Tucker decomposition. For example, if we wish to transfer in a set of existing roles, we can adjust Algorithm 1 and not solve for the $F$ matrix that defines the roles. This allows us to test many interesting questions such as how transferable the roles from other graphs are at explaining another multi-relational graph. We exper-iment with this particular type of transfer in Figure 18, however other types of transfer are possible. We now discuss all types but due to space limitations show experiments only for role transfer.

**Role transfer** can be used to detect to what extent roles are similar or dissimilar across different multi-relational graph. If there is a particularly interesting set of roles that have been studied in another graph, they can be transferred to a new graph to see how the nodes in that graph play those roles.

**E-group transfer** can only be used if the multi-relational graphs are on the same entities. However if there are some well understood grouping of entities (say Democrat, Tea Party and Republican) these can be translated into E-groups and transferred to help gain understanding of the behaviors of those specific groups.

**R-group transfer**, similar to role transfer, can be used to test how well relation groupings transfer across multiple graphs.

## 9 Empirical Results

As in our previous work, all code and data sets will be made publicly available on our website.

Since we wished to focus on analyzing both multi-relational graphs and collections of similar multi-relational graphs for transfer setting, we focused our empirical analysis on the Cosponsorship Network Data [15, 16] data set. This data set consists of congressional cosponsor data for over 30 years of congresses. Congressional representatives have the ability to add their name to a bill in order to lend support to it (called cosponsoring), and it has been argued that this act is a good measure of interaction within congress because legislators spend considerable effort convincing other representatives to cosponsor their bills. Using this publicly available information about cosponsorships, each congress can be broken up into a multi-relational graph with approximately 450 different nodes (congressional representatives) who jointly cosponsor approximately 10,000 bills per congress (many are just amendments). Table 6 show statistics for the graph created from the 110th congress, but in all we study the 96th-110th congresses, each of which has their own cosponsorship graph. Rather than create a cosponsorship graph based on all of the proposed bills from a particular congress, we build a multi-relational graph by viewing each committee as a separate relation (see Figure 10). Each bill is assigned to a committee based on the topic of the legislation. We analyzed bills from 15 different committees (the committees for which there were legislation in each congress 96th-110th) so that all of the relations are consistent over all the multi-relational graphs. Across the different congresses the one factor that does change is the set of elected representatives elected during each. Putting this altogether the multi-relational graph we study is a $person \times person \times committee$ tensor such that the entry at $(i, j, k)$ indicates how often congressman $i$ and $j$ **cosponsored** a bill that was sent to committee $k$ for a particular congress. This graph has many underlying complexities in terms of groups of congressional representatives who work together (i.e., party-based and tenure-length based), the roles that congressional representatives play (e.g., focused and generalist), and the relationships of the various bill areas (e.g., science-focused, business-focused). **We study the last 15 congresses (96th to 110th) and have a multi-relational graph for each.**

| Graph Attribute | Value |
|---|---|
| Number Representatives | 453 |
| Number Bills | 10613 |
| Sponsors Per Bill | 16.9 |
| Mean cosponsor degree (aggregated) | 8.37 |
| Standard deviation (aggregated) | 6.31 |
| Number of zeros (aggregated) | 1729 |
| Mean cosponsor degree (median) | 0.48 |
| Standard deviation (median) | 1.02 |
| Number of zeros (median) | 53235 |

Table 6: Details on the congressional cosponsor data set for the 110th congress. The aggregated statistics were calculated on the cosponsorship graph without treating it as a multi-relational graph. The median statistics measure the median attribute value over each relation or committee. The number of zeros refers to the number of pairs of representatives that have no edge (or an edge of weight zero).

## 9.1 Studies on a Single Multi-Relational Graph

Here we present results on the analysis of the 110th Congress which sat from 2007-2009. This was a Democrat controlled congress that sat during the last two years of President George W. Bush's administration. It was also unique in that it was the first Democrat controlled congress since 1995.

We analyzed this multi-relational cosponsor graph using our formulation for multi-relational role discovery. This produced E-groups, roles, and R-groups along with an interaction graph that explained the interactions between the three concepts. The E-groups are shown in Figure 12, the interpretation of role definitions is shown in Figure 11, and the composition of the R-groups is shown in Figure 13. How these E-groups, roles, and R-groups interact in the interaction graph are shown both directly as a sliced core in Figure 14, as a sparsified graph in Figure 15, and as a graph embedding in Figure 16.

**Underlying E-groups, Roles and R-groups.** Figure 12 shows that as expected people from the same party cosponsor the same bills though this further divides into two different E-groups per party. For the two Democrat groups we note that there is an E-group of mostly junior congressmen (group 4) whilst the other contains many of the senior congresswoman (group 1). Of particular note is the 5th E-group that contains a mix of Republican and Democrat representatives which largely represents a group of centralist members. For example McGotter was a well known member of the moderate "Republican Main Street

Partnership".

Figure 11 shows the types of roles that are found in the graph via sense making [21]. This plot shows for each role the attributes shared by representatives who play that role. Roles can be contrasted and compared in terms of these reference features. For example roles 2 and 4 both have comparable degree but largely differing weight, meaning representatives from both roles participated in cosponsorship with roughly the same number of other representatives, but representatives in role 4 cosponsored with the same people more often.

Figure 13 shows the compositions of the R-groups. Each R-group is composed of some combination of the 15 studied relations each of which in turn correspond to a congressional committees which is roughly interpretable as the topic of the bill. While there is some overlap in the relational contribution of each R-group, each of them has a unique dominating relation (R-group 1 'Ways and Means', R-group 2 'Rules', R-group 3 'Oversight and Government Reform', R-group 4 'Education and Labor', R-group 5 'Agriculture'). Because we did not enforce orthogonality for our Tucker decomposition, as is commonly done (see Algorithm 1), we can see which relations are less distinguishing in terms of role analysis by looking at those relations that show up in multiple R-groups (e.g., 'Transport and Infrastructure' is assigned to every R-group).

**Interactions Between E-groups, Roles and R-groups.** We now explain the Interaction Graph which is shown in Figure 15. As previously mentioned E-groups are largely divided by party even though party was not part of the data set. It can be argued then that this role discovery formulation discovered communities rather than roles. However the reason these groups divided along party lines is because parties are playing different roles in different R-groups. Depending on different factors such as which party is the majority, we expect the parties to play different roles, so our analysis matches our expectations.

While there is much overlap in the R-groups that both parties participate in, the parties play different roles in those R-groups. For example the Republican groups participate largely in R-groups 3,4,5 while the Democrat groups participate largely in R-groups 1,2,3,5. However E-group 4 (Republican) and E-group 5 (Democrat) play different roles in R-group 5 (Agriculture). This is an example of a *Role Tie* from Figure 9.

There are also some roles and E-groups that are unique to a party. For example role 2 is exclusive to Republicans (many collaborators, but not many collaborations). And R-group 1 (Ways and Means) is more strongly associated with the Democrat E-groups. This makes sense, because the Ways and Means committee is one of the most prestigious to participate in and relates to tax legislation. It therefore makes sense that the majority party would be most active in this committee.

Though the direct view of the interaction graph is useful, as discussed earlier there are other methods to understand the interaction. We can slice the core tensor either by E-group, role, or R-group and directly compare. Figure 14 shows such a comparison across E-groups. We can see that E-groups 1 and 3 both play role 5 but on different R-groups, also E-group 1 plays mainly one role, but E-group 3 plays multiple roles in the graph. Finally, we can embed this graph into a metric space as shown in Figure 15.

## 9.2 Studies Across Multiple Multi-Relational Graphs

We also performed multi-relational analysis across a total of 15 consecutive congresses and report the results here. There were two experiments we performed, to analyze these multi-relational graphs and to gain insight into them. First in Figure 17 we analyzed how the macro-properties of the learned interaction graphs, as discussed in Section 8.3, varied throughout the congress (see Figure 17). And second we determined how well roles definitions learned from one congress can transfer to others, as discussed in Section 8.5, the results of which are presented in Figure 18.

Figure 17 shows the results of our analysis of macro-properties of the learned interaction graphs from the 96th-110th congresses. These results contain an immense amount of interesting insights and we focus on just a few due to space restrictions. The first unusual property is we note is a great spike of instability in the 101st congress. This is due to the election of a new President Bush following a very popular bipartisan President Regan. In addition many controversial bills were passed that crossed party lines such as the Americans With Disabilities Act. In contrast the 99th congress was very stable given it was Regan's second term and most bills were supported across partisan lines. Of particular note is also the sharp peaks during congresses 97, 101 and to lesser extent 103. They correspond precisely to changes in Presidencies: Carter (Democrat) to Regan (Republican) (97), Regan (Republican) to Bush (Republican) (101) and Bush (Republican) to Clinton (Democrat) (103).

In Figure 18 we show a heat map on the role transfer between different congresses. We first ran our algorithm to discover the roles for all congresses. Then we transferred each set of role definitions learned from all 15 congresses to every other congress, and measured the fit to determine how well each set of roles could be used to explain the behavior of every other congress. The heat map shows how well (dark red) or how poorly (dark blue) the roles for the congress in the row explained the interactions for the congress in the column. Of course the diagonal is dark red since those roles were built from data for that congress. As expected the block red structure indicates that later congresses roles can better explain later congresses behavior and earlier congresses roles can explain earlier congresses behavior. The solid blue block on the top left hand corner indicates that later congresses roles are very poor at explaining the later congresses behavior. The apparent outliers within the top right hand block and lower left hand block (i.e., the bluish entries amongst the red/yellow) are indicative of a shift in presidency or house majority either Democrat to Republican or vice-versa.

# 10   Conclusion

Role discovery is an emerging and important area of graph mining. It looks at discovering nodes that perform similar functions in networks, but do not necessarily belong to the same community. Existing work so far has had two limitations: they are completely unsupervised and are focused on single relational graphs.

We propose a framework that allows incorporating convex constraints into NMF to allow a rich set of guided role discovery formulations. In particular we explore three types of guidance: sparsity, diversity and alternativeness. Sparsity and diversity can be used to create simpler and more interpretable role definitions and role allocations. Also they can reduce overfitting and produce better predictive results for matching authors between the KDD conference and a variety of other conferences provided they perform similar roles in both conferences. The notion of alternativeness has been explored in the clustering literature and is useful if the given explanation is not valid and an alternative is required. Here we show that not only do alternative roles exist in co-author networks, but that we can find an alternative to the community structure in a very large YouTube graph.

We then showed how to lift that framework to multi-relational graphs by first representing the multi-relational graph as a tensor. We then use a Tucker decomposition due to the more popular PARAFAC decomposition not being able to find the complex interactions that are likely to occur between the E-groups, roles, and R-groups. However, existing Tucker decomposition algorithms in popular toolboxes enforce properties that would lead to non-intuitive results for role discovery, hence we formulate our own algorithm. A critical aspect to our work is how to interpret and use the core of the Tucker decomposition which shows the complex interactions between the E-groups, roles, and R-groups. We show how it can be visualized and represented as an interaction graph whose properties we can use as macroscopic indicators of the original multi-relational graph. Our experimental results focus on 15 multi-relational Congressional cosponsor record graphs. Here an E-group is a collection of congressional representatives, an R-groups is the collection of bill types (determined by the committee they went through), with the roles being on cosponsoring behavior. We show that our methods can find intuitive and expected insights such as Republican and Democrats naturally separate into different E-groups. We also find that groups of representatives can play multiple roles for multiple R-groups, showing that the Tucker decomposition does indeed find the complex interactions we wish to discover. The macroscopic properties of the interaction graph show that the congresses vary greatly over time with abrupt changes being associated with changes in the Presidency and control of the Congress. Finally our transfer setting offers a useful insight into understanding how roles have differed across congress by using the roles from different congresses to explain the behavior of others.

## 11 Acknowledgments

The authors gratefully acknowledge support of this research via ONR grants N00014-09-1-0712, N00014-11- 1-0108 and NSF Grant NSF IIS-0801528. This work was also supported in part by IARPA via AFRL Contract No. FA8650-10-C-7061 and in part by DAPRA under SMISC Program Agreement No. W911NF-12-C-0028.

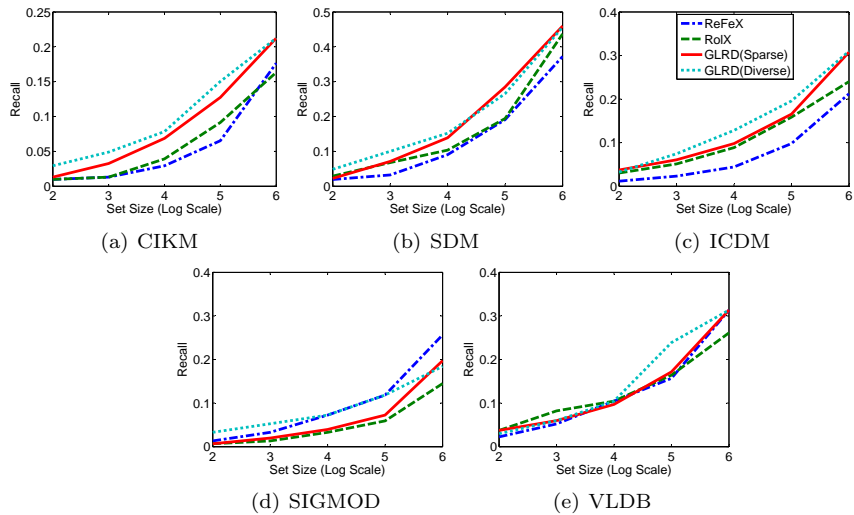

Figure 3: Comparison of role discovery techniques for identity resolution across graphs. Role definitions are learned from the KDD co-authorship graph; then, authors from the other (conference) co-authorship graphs are assigned to these roles using various techniques. In particular, we show results for ReFeX (features only), RolX (unconstrained role discovery), GLRD-Sparse (role discovery with sparsity constraints), and GLRD-Diverse (role discovery with diversity constraints). Authors from each conference are paired with increasing number of nearest neighbors from KDD conference (x-axis) and the resulting recall is reported (y-axis). Across most settings role definitions using sparsity and diversity constraints lead to better identity resolution results than standard unconstrained RolX. For graphs that are most similar in nature to KDD (e.g. ICDM, SDM, CIKM) the transfer of role definitions lead to better results than simply using structural features of nodes directly. Note that the recall values are relatively low because the set sizes (on the x-axis) are small compared to the population size in each graph.

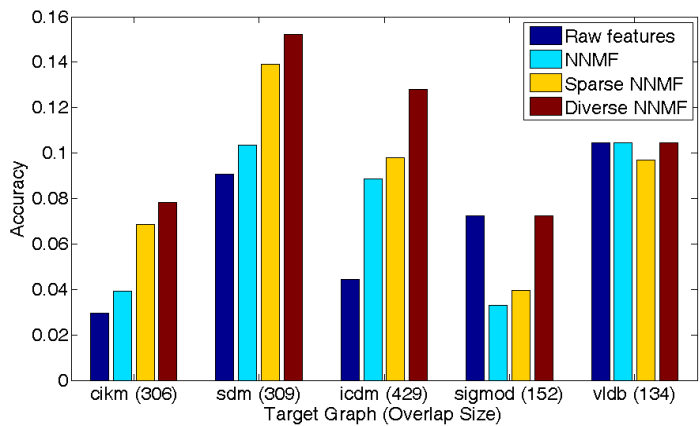

Figure 4: Comparison of role discovery techniques for identity resolution experiments. Authors from each conference paired with the nearest 32 neighbors from KDD conference; the resulting recall accuracy is reported. The percentage number (on the x-axis) is the fraction of authors that overlap between the two conferences. Nearly all experiments show better results with sparsity and diversity constraints except when the authors do not share similar roles in the two conferences (SIGMOD and VLDB).

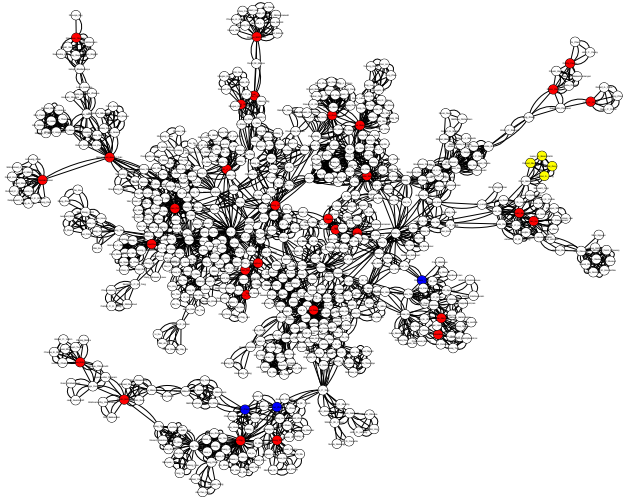

Figure 5: A visualization of our alternative role discovery results for the KDD co-authorship graph's largest connected component. All the colored nodes belong to the same primary role under the original factorization. However, they belong to different primary roles under the alternative factorization, as indicated by the various colors. We observe that the alternative roles are able to separate the 3 blue "local-star" nodes (namely, Jun Zhu, Lei Zhang, and Evimaria Terzi) from the red "global-broker" nodes (namely, Christos Faloutsos, Heikki Mannila, Vipin Kumar, etc). The alternative roles also separate out the 4 yellow "periphery-cliquey" nodes. Note, the reader can zoom in on this figure to read the names of each author.

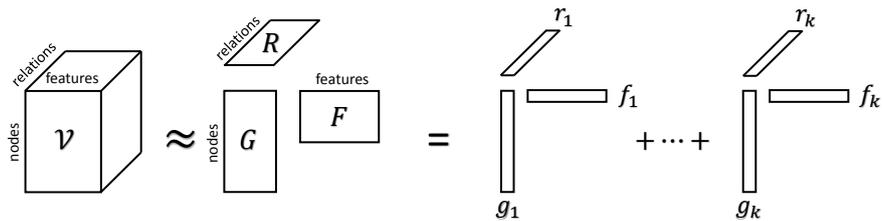

Figure 6: A multi-relational graph represented using an order 3 tensor. The PARAFAC tensor decomposition is a rank 1 simplification of the graph and is the natural analog to the earlier used [21, 18] NMF formulation of role discovery. However, it has significant limitations for role discovery in multi-relational graphs.

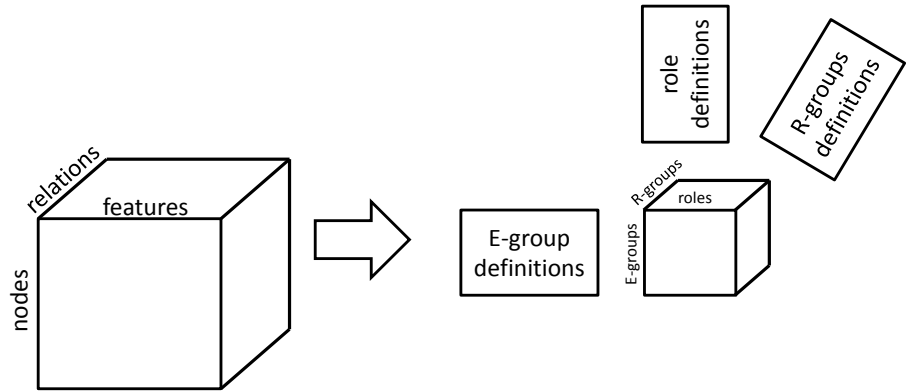

Figure 7: The Tucker decomposition for role discovery. The factor matrices can be interpreted as: groups of features (role definitions), groups of entities (E-groups), and groups of relations (R-groups). The Tucker core shows how the roles/E-groups/R-groups interact in the multi-relational graph and can be viewed itself as a hyper-graph which we call an example of an interaction graph.

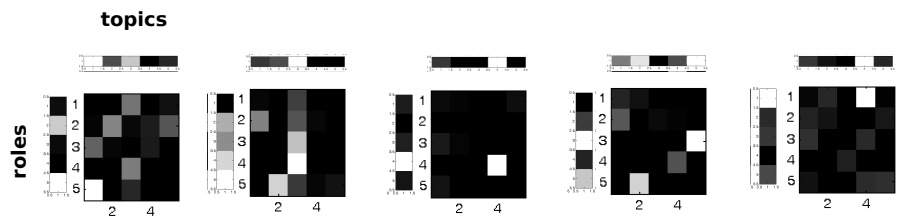

Figure 8: Analysis of E-group slices from the tensor core. Each slide shows the roles/R-groups each E-group of people play and are directly comparable.

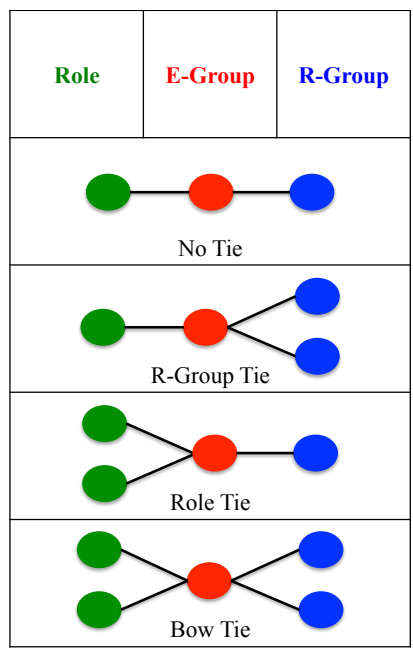

Figure 9: Some patterns that can exist in an interaction graph. *No Tie*: E-group only plays one role in one R-group; *R-Group Tie*: E-group plays same role in multiple R-groups; *Role Tie*: E-group plays multiple roles in same R-groups; *Bow Tie*: E-group plays multiple roles but in different R-groups.

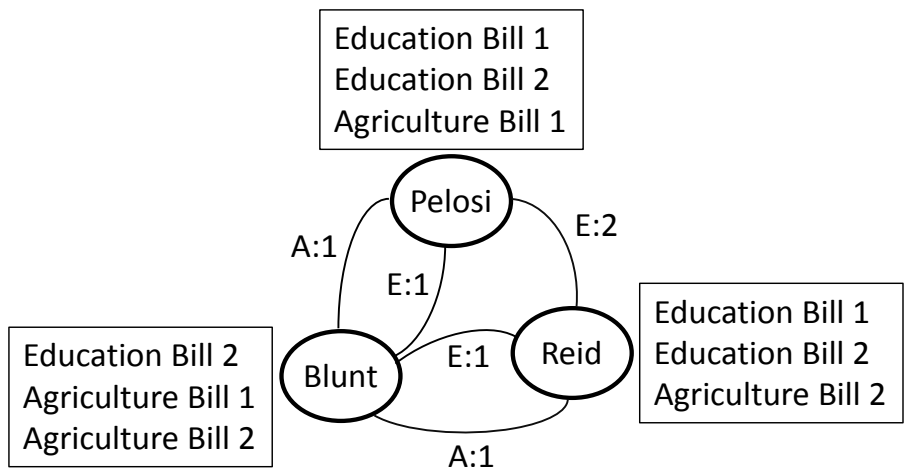

Figure 10: Description of how multi-relational graphs are created from the congressional cosponsors data. Nodes in this graph represent congressional representatives and the adjacent lists of hypothetical bills are those that the representative cosponsored. When two representatives cosponsor the same bill, a labeled edge is created between them where the label corresponds to the assigned committee for the bill (e.g. Agriculture, Education). The weight associated with a labeled edge corresponds to the number of bills from the same committee a pair of representatives both cosponsored.

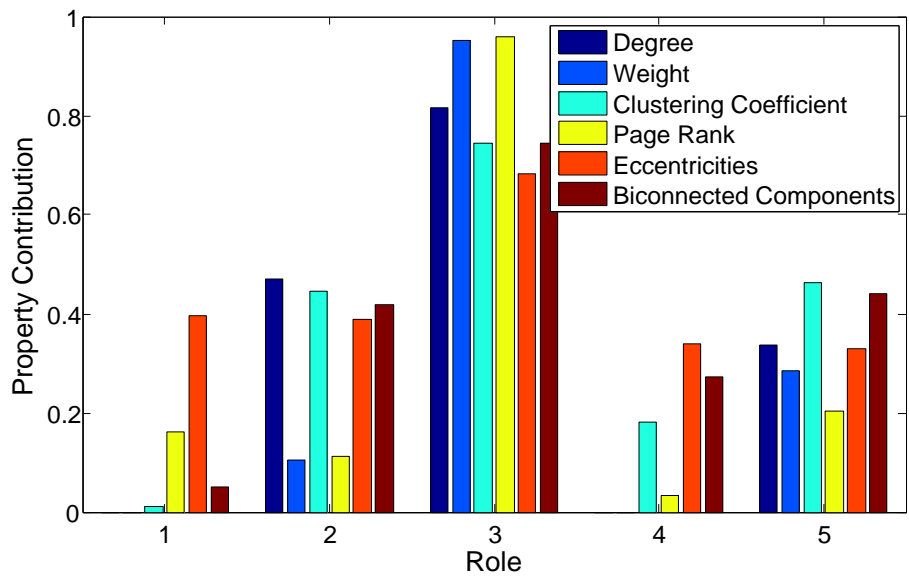

Figure 11: Sense making of roles discovered in the 110th Congress Cosponsor Multi-Relational Graph. Roles are redefined in terms of a set of reference features each of which is normalized for comparison purposes. Role 3 are the power brokers.

| E-group 1 | | |
|---|---|---|
| Name | Party | Exp |
| Millender-McDonald | D | 11 |
| Obey, David | D | 38 |
| Tsongas, Niki | D | 0 |
| Speier, Jackie | D | 0 |
| Faleomavaega, Eni | D | 18 |
| Meehan, Martin | D | 14 |
| Edwards, Donna | D | 0 |
| Visclosky, Peter | D | 22 |
| Hoyer, Steny | D | 26 |
| Foster, Bill | D | 0 |

(a) Democrat seniority. Hoyer was the majority leader. Characterized by large number of collaboration with many representatives largely in 3rd R-group (Ways and Means).

| E-group 2 | | |
|---|---|---|
| Name | Party | Exp |
| Hensarling, Jeb | R | 4 |
| Boehner, John | R | 16 |
| Thornberry, Mac | R | 12 |
| Broun, Paul | R | 0 |
| Shadegg, John | R | 12 |
| Hastert, Dennis | R | 8 |
| Scalise, Steve | R | 11 |
| Latta, Robert | R | 6 |
| Flake, Jeff | R | 6 |
| McCrery, Jim | R | 14 |

(b) Republican seniority. Boehner was minority leader at the time.

| E-group 3 | | |
|---|---|---|
| Name | Party | Exp |
| Cooper, Jim | D | 16 |
| Johnson, Henry | D | 0 |
| Ryan, Tim | D | 4 |
| DeGette, Diana | D | 10 |
| Engel, Eliot L. | D | 14 |
| Doggett, Lloyd | D | 12 |
| Pastor, Ed | D | 16 |
| Meek, Kendrick | D | 4 |
| Murphy, C. | D | 0 |
| Crowley, Joseph | D | 8 |

(c) Active largely in R-group (5th) but with multiple roles. The 5th R-group is dominated by the agriculture committee.

| E-group 4 | | |
|---|---|---|
| Name | Party | Exp |
| Hall, Ralph | R | 16 |
| Rodgers, Cathy | R | 2 |
| Myrick, Sue | R | 12 |
| Issa, Darrell | R | 6 |
| Drake, Thelma | R | 2 |
| Kuhl, Randy | R | 2 |
| Poe, Ted | R | 2 |
| Boozman, John | R | 6 |
| Conaway, Michael | R | 2 |
| Wamp, Zach | R | 12 |

(d) Working with many representatives (high degree) but not often (low weight) on R-group 5.

| E-group 5 | | |
|---|---|---|
| Name | Party | Exp |
| Jackson-Lee, Sheila | D | 12 |
| Cohen, Steve | D | 0 |
| Hare, Phil | D | 0 |
| Grijalva, Raul | D | 4 |
| English, Phil | R | 12 |
| Honda, Michael | D | 6 |
| McCotter, Thaddeus | R | 4 |
| Filner, Bob | D | 14 |
| Hinchey, Maurice | D | 14 |
| Gonzalez, Charles | D | 8 |

(e) Mixed party membership

Figure 12: Samples of congressional representatives from each E-group (found in in the 110th Congress Cosponsorship Graph) along with their party affiliation and years of service in U.S. House of Representatives at beginning of congress (2007).

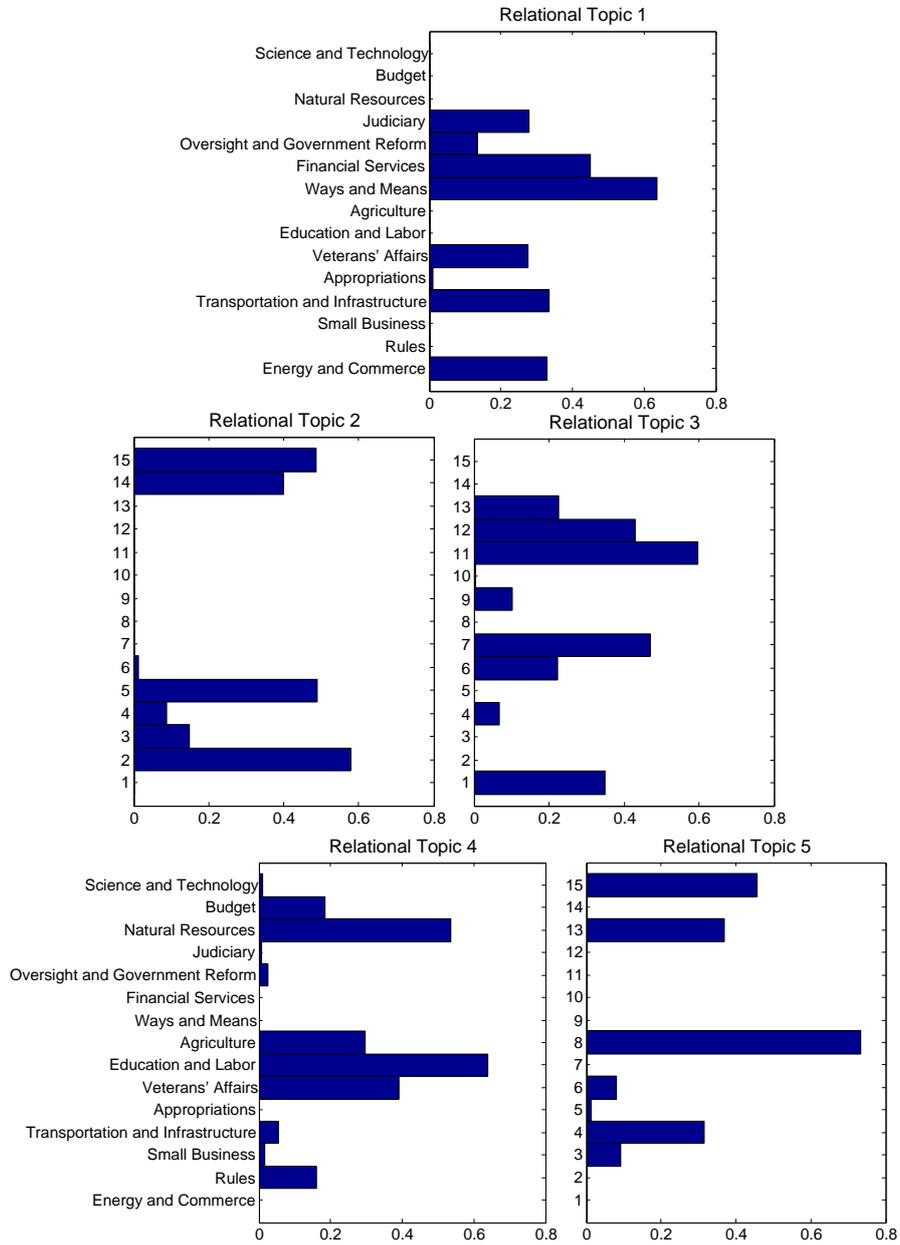

Figure 13: R-groups for 100th congress. Each bar plot corresponds to a single R-group and the bars show how much each relation contributes to the respective relation R-group.

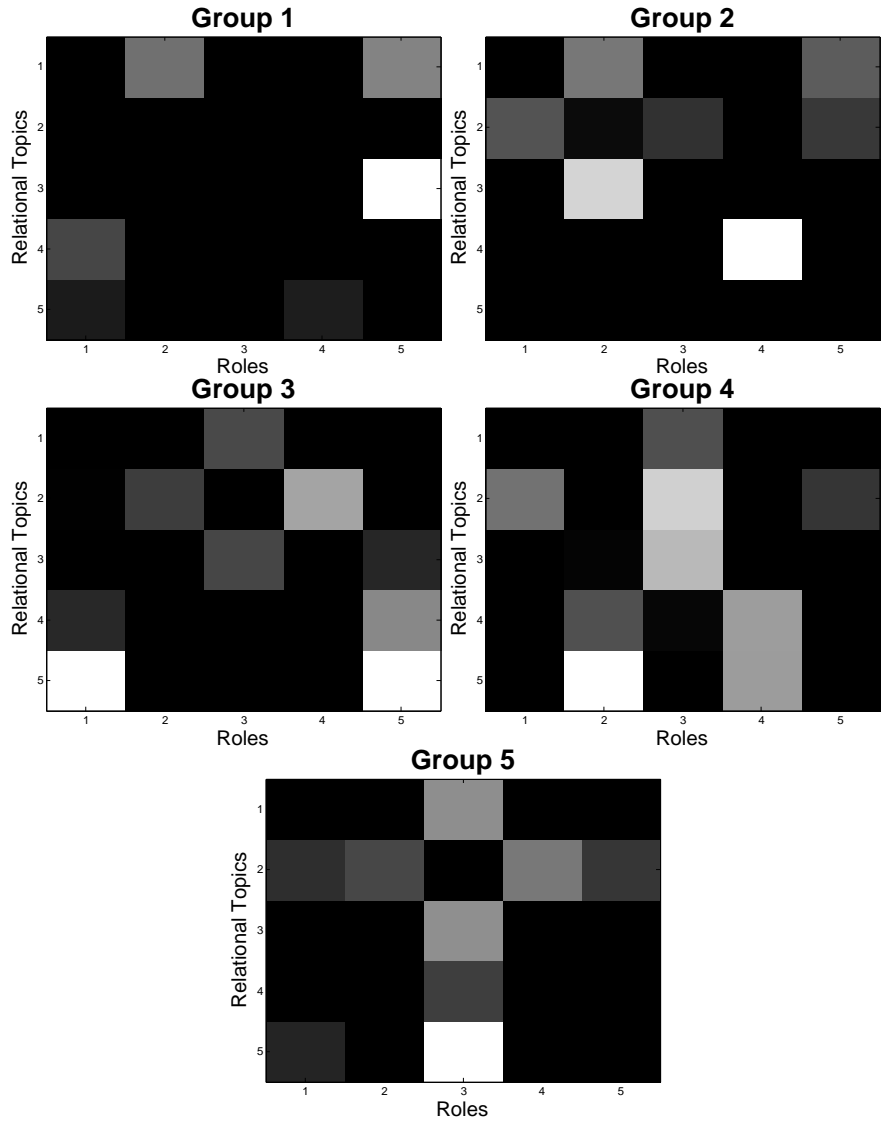

Figure 14: Tucker core found in in the 110th Congress Cosponsorship Graph sliced by E-group. Each slice represents an E-group while the rows correspond to R-groups and the columns correspond to roles. Light colors correspond to high values and black corresponds to zero value.

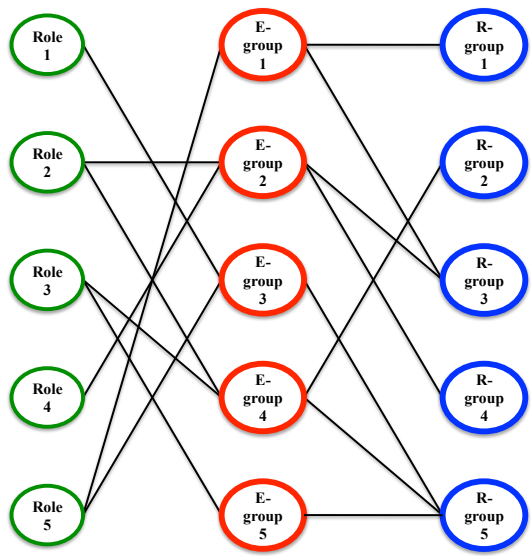

Figure 15: Sparsified tripartite representation of core tensor found in the 110th Congress Cosponsorship Graph. Each entry $i, j, k$ of core corresponds to a hyperedge between E-group $i$, role $j$, and R-group $k$ We sparsify this into a single-relation graph that is role focused. Looking back to our example patterns (see Figure 9), we observe that this congress has two *bow ties* patterns (E-groups 2 and 4, 100% Republicans); one *no tie* pattern (E-group 5, 80% Democrats), one *role tie* pattern (E-group 3, 100% Democrats), and one *R-group tie* (E-group 1, 100% Democrats). Figure 12 lists the members of each E-group.

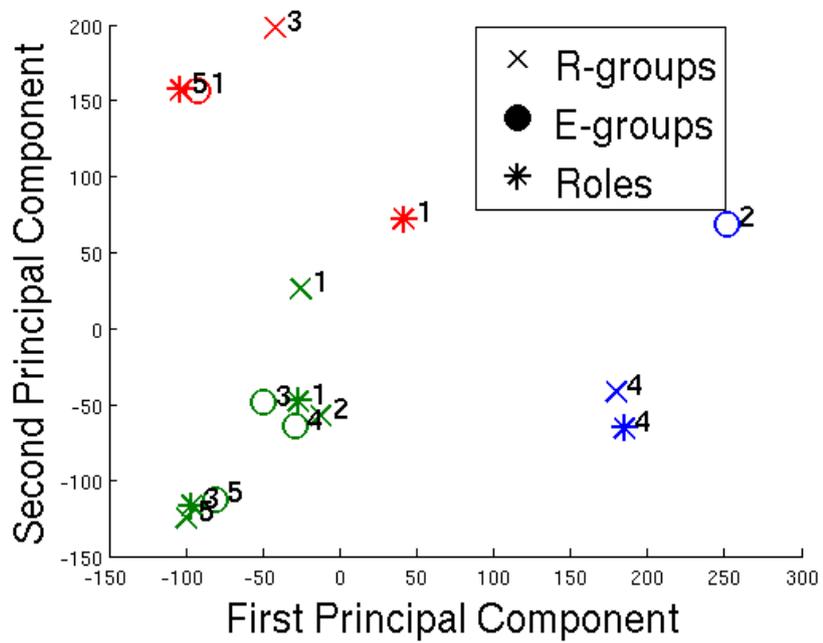

Figure 16: Projection and heterogeneous clustering of tripartite graph representation of core tensor found for the 110th Congress Cosponsor Graph. Colors represent the clustering while marker shapes represent the type of object (E-groups, roles, and R-groups).

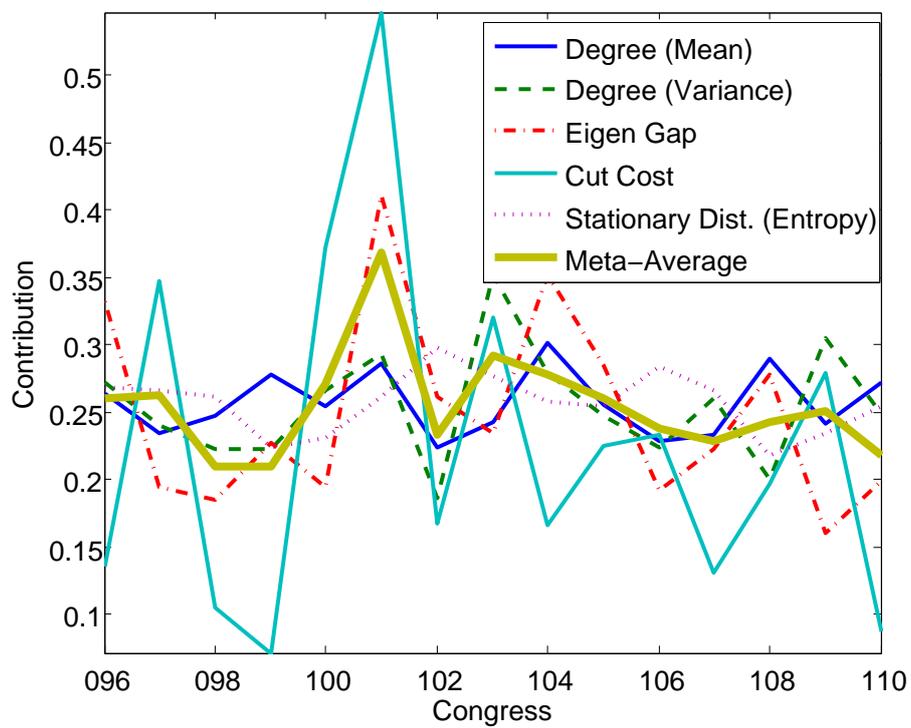

Figure 17: Properties of interaction graph formed from the Tucker cores for the last 15 Congresses. Attributes are all normalized for comparison purposes.

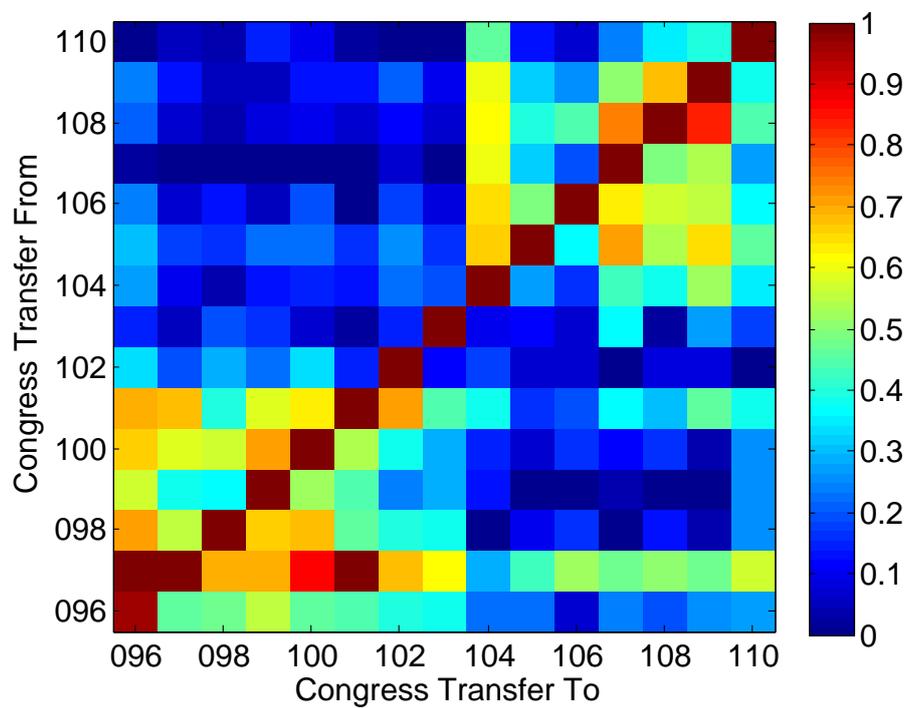

Figure 18: Fit quality when transferring roles between 96th to 110th congress. $fit = 1 - reconstruction\ error/||\mathcal{V}||$. Roles learned from congresses on the x-axis are transferred to each congress as denoted on the y-axis. Transferring to temporally further congresses generally leads to poorer fits.